\documentclass[11pt]{article}
\usepackage[margin=1in]{geometry}
\usepackage{graphicx}
\usepackage{amsmath}
\usepackage{amssymb}
\usepackage{natbib}
\usepackage{hyperref}
\usepackage{caption}
\usepackage{subcaption}
\usepackage{booktabs} 
\usepackage{siunitx}  
\usepackage{hyperref}

\title{\textbf{PUNCH: Physics-informed Uncertainty-aware Network for Coronary Hemodynamics}}

\author{
Sukirt Thakur$^{1}$, Marcus Roper$^{2}$, Yang Zhou$^{1}$, Dmitry Yu. Isaev$^{1}$, \\
Reza Akbarian Bafghi$^{3}$, 
Brahmajee K. Nallamothu$^{4}$,  C. Alberto Figueroa$^{5,6}$,  \\ Srinivas Paruchuri$^{1}$, Scott Burger$^{1}$, 
 Carlos Collet$^{7}$, Maziar Raissi$^{8}$ \\
\\
\scriptsize
$^{1}$AngioInsight Inc., USA \\
\scriptsize
$^{2}$Depts. of Mathematics and Computational Medicine, University of California, Los Angeles , Los Angeles, CA 90095, USA \\
\scriptsize
$^{3}$University of Colorado Boulder, Boulder, CO 80309, USA \\
\scriptsize
$^{4}$Department of Internal Medicine, University of Michigan, Ann Arbor, MI 48109, USA \\
\scriptsize
$^{5}$Department of Biomedical Engineering, University of Michigan, 48109, Ann Arbor, MI, USA \\
\scriptsize
$^{6}$Department of Surgery, University of Michigan, 48109, Ann Arbor, MI, USA \\
\scriptsize
$^{7}$
Clinical Trial Center, Cardiovascular Research Foundation, New York, NY, United States \\
\scriptsize
$^{8}$University of California, Riverside, Riverside, CA 92521, USA \\
\\
\scriptsize
\textit{Correspondence to:} Sukirt Thakur (sukirt@angioinsight.com)
}

\date{}
\begin{document}

\maketitle

\begin{abstract}
More than 10 million coronary angiograms are performed globally each year, providing a gold standard for detecting obstructive coronary artery disease. However, no obstructive lesions are identified in 70\% of patients evaluated for ischemic heart disease. Up to half of these patients have undiagnosed life-limiting coronary microvascular dysfunction (CMD), which remains under-detected due to the limited availability of invasive tools required to measure coronary flow reserve (CFR). Here, we introduce PUNCH, a non-invasive, uncertainty-aware framework for estimating CFR directly from standard coronary angiography. PUNCH integrates physics-informed neural networks with variational inference to infer coronary blood flow from first-principles models of contrast transport, without requiring ground-truth flow measurements or population-level training. The pipeline runs in approximately three minutes per patient on a single GPU. Validated on synthetic angiograms with controlled noise and imaging artifacts, as well as on clinical bolus thermodilution data from 20 patients, PUNCH demonstrates accurate and uncertainty-calibrated CFR estimation. This approach establishes a new paradigm for CMD diagnosis and illustrates how physics-informed inference can substantially expand the diagnostic utility of available clinical imaging.
\end{abstract}

\section*{Introduction}
Coronary microvascular dysfunction (CMD) is a prevalent, yet underdiagnosed condition that affects up to 70\% of patients who have symptoms of angina but no detectable obstruction in their coronary arteries \cite{kunadianEAPCIExpertConsensus2021}. The diagnostic gap disproportionately impacts women \cite{ongInternationalStandardizationDiagnostic2018, WISE} and low income patients \cite{spetko_distance_2025}. Unlike epicardial stenosis, CMD reflects abnormalities in flow and structure that are invisible on standard angiography but critically influence myocardial perfusion and patient outcomes \cite{murthyCoronaryVascularDysfunction2012, taquetiCoronaryMicrovascularDisease2018}.

The gold standard for diagnosing coronary microvascular function is the measurement of coronary flow reserve (CFR) using invasive catheter-based techniques such as Doppler flow wires or thermodilution \cite{taquetiCoronaryMicrovascularDisease2018,barbatoValidationCoronaryFlow2004}. In thermodilution, a known volume of cold saline is injected into the coronary artery, and downstream temperature changes are recorded by a thermistor-tipped catheter. The resulting temperature--time curve—an indicator-dilution profile—allows for estimation of blood flow based on the area under the curve \cite{fearonNovelIndexInvasively2003}. Although reliable, thermodilution requires precise timing, is sensitive to injection technique and temperature variability, and carries procedural risks inherent to placement of a coronary guide-wire in the distal coronary artery \cite{debruyneCoronaryFlowReserve1994, fearonNovelIndexInvasively2003}. These practical constraints have limited the widespread invasive evaluation of CFR, leaving most patients without a definitive functional diagnosis.

\begin{figure}[!h]
    \centering
    \includegraphics[width=\linewidth]{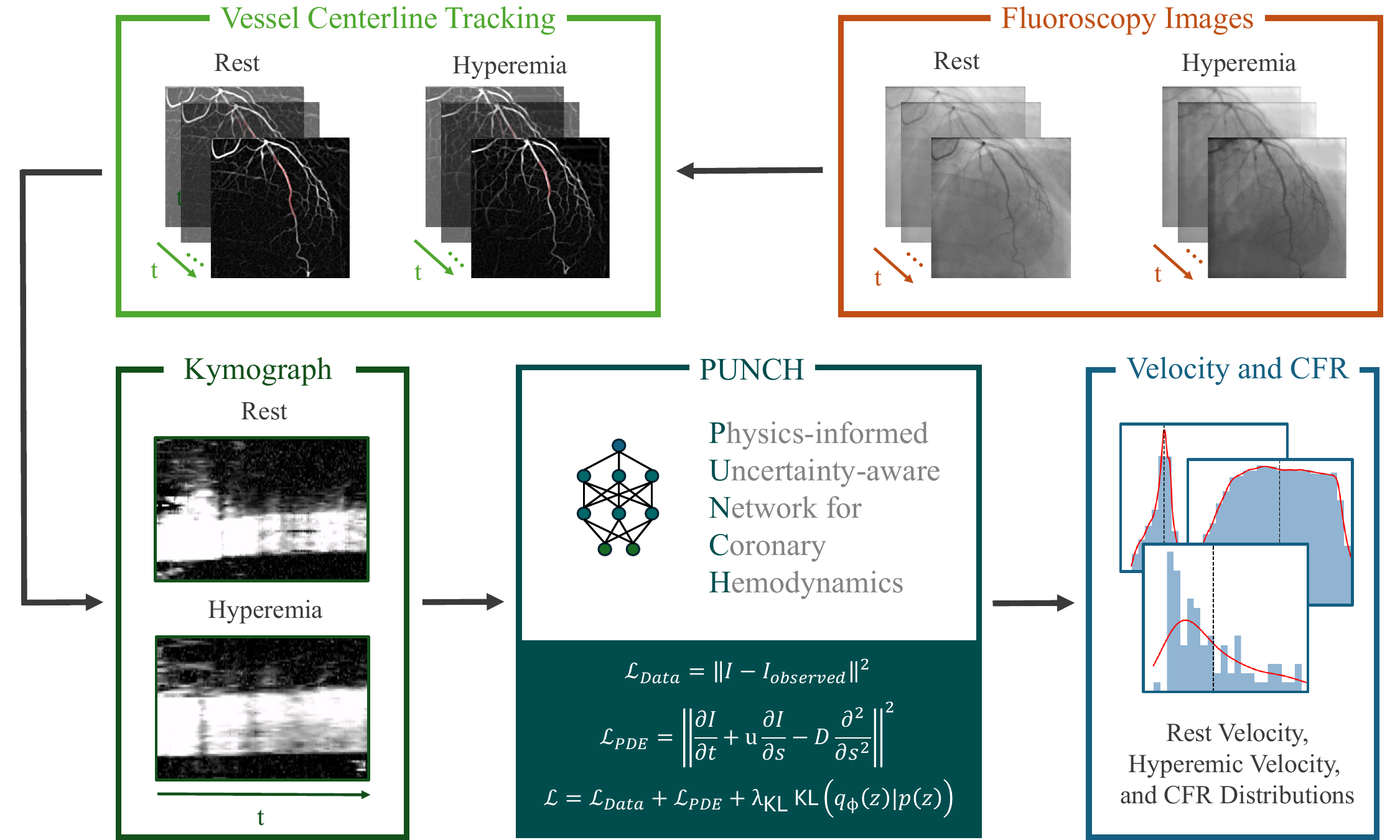}
    \caption{\textbf{PUNCH enables uncertainty-aware estimation of CFR from standard angiography.} Fluoroscopy images acquired at rest and hyperemia undergo vessel centerline tracking to generate spatiotemporal intensity maps (kymographs). These kymographs serve as input to a physics-informed neural network that jointly optimizes data fidelity ($\mathcal{L}_{Data}$), advection-diffusion physics constraints ($\mathcal{L}_{PDE}$), and variational inference regularization ($\mathcal{L}_{KL}$). The framework outputs posterior distributions of rest velocity, hyperemic velocity, and CFR, providing probabilistic estimates with quantified uncertainty. The dual-branch architecture models both physiological states while sharing a common latent variable z to capture measurement uncertainty from noise, motion artifacts, and incomplete contrast filling.}
    \label{fig:PUNCH_pipeline}
\end{figure}

In principle, contrast-enhanced coronary angiography encodes flow information analogous to thermodilution: the transport and dispersion of contrast dye through the coronary lumen reflect local hemodynamics \cite{zierlerIndicatorDilutionMethods2000, ballFractionalFlowReserve2018}. Recently, there has been an interest in using coronary angiography as a diagnostic tool for coronary microvascular dysfunction using data-driven approaches\cite{yangAssessingCoronaryMicrovascular2026}. However, extracting quantitative flow metrics from these spatiotemporal image patterns remains challenging due to image noise, cardiac motion, vessel overlap, and the absence of direct velocity measurements.

Recent advances in physics-informed machine learning offer a promising pathway for addressing the inverse problem of inferring `hidden' hemodynamic variables, such as blood velocity, from observed contrast dye transport dynamics. In contrast to conventional data-driven approaches, physics-informed neural networks (PINNs) incorporate known physical laws directly into the learning process, allowing the estimation of hidden physiological quantities from indirect measurements, such as medical images \cite{raissiPhysicsinformedNeuralNetworks2019}. However, most existing PINN-based approaches yield only deterministic point estimates and provide no measure of confidence, which is an important limitation in clinical settings where image quality, motion, and contrast delivery can vary substantially \cite{yangAdversarialUncertaintyQuantification2019, zouUncertaintyQuantificationNoisy2025, linkaBayesianPhysicsInformed2022}.

Here, we introduce an uncertainty-aware, physics-informed framework to estimate CFR directly from standard coronary angiography. The key insight is simple: the contrast material injected during angiography behaves as an intravascular tracer. As it travels through the coronary artery, it is transported by blood flow and simultaneously spreads due to velocity heterogeneity and mixing. These observable spatiotemporal patterns encode information about coronary blood flow and microvascular resistance.

Our approach leverages this physical behavior to infer flow dynamics from angiographic image sequences while explicitly quantifying uncertainty. Rather than relying on population-level training or labeled flow data, each patient is independently analyzed using first-principles contrast transport physics. Importantly, the framework produces not only a CFR estimate, but also a confidence interval that reflects the uncertainty related to image quality, motion artifacts, and incomplete contrast filling. This uncertainty-aware formulation enables a safer interpretation of angiography-derived physiology and provides a pathway toward robust, non-invasive assessment of coronary microvascular function during routine clinical care.

\begin{figure}[!h]
    \centering
    \includegraphics[width=\linewidth]{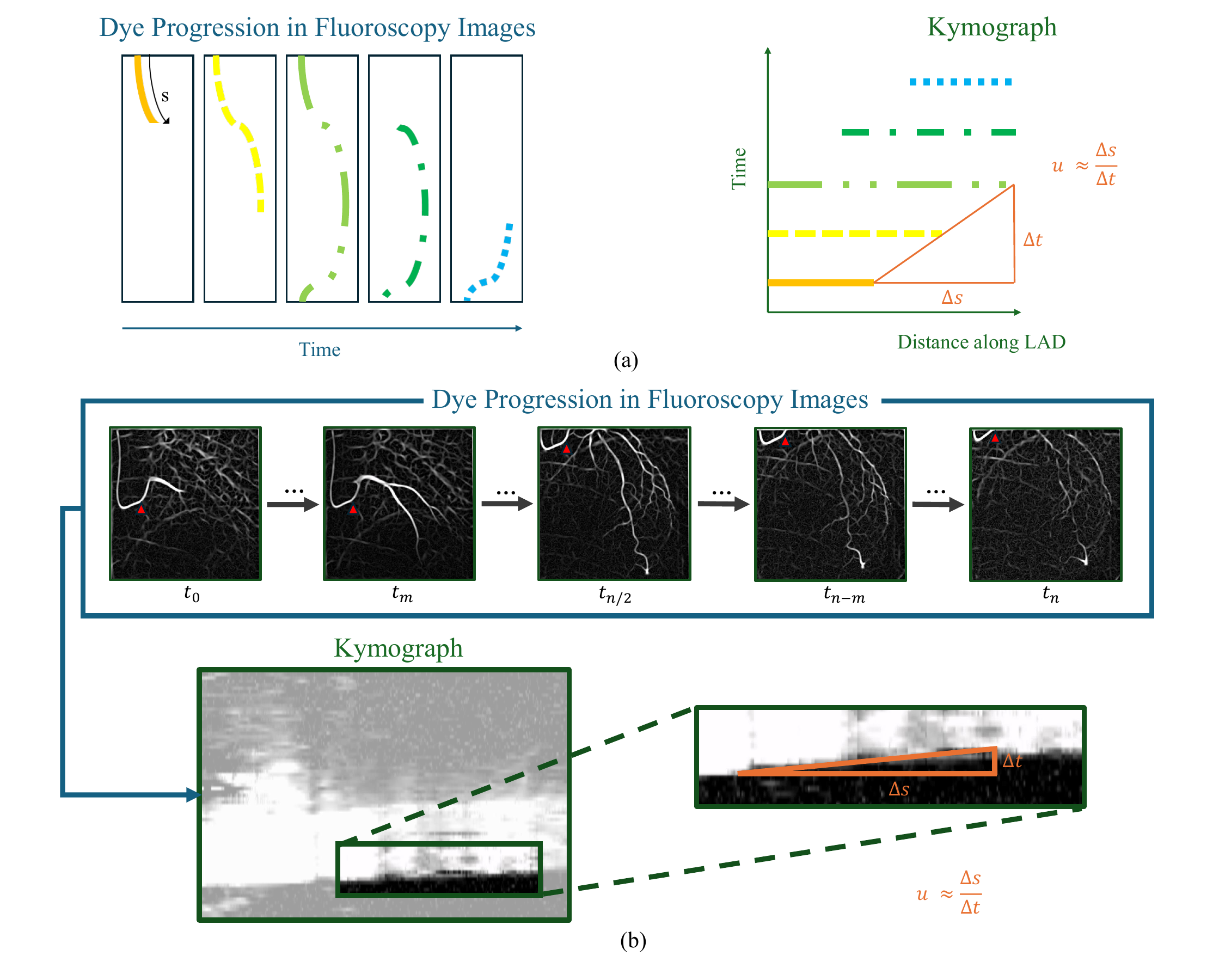}
    \caption{\textbf{Kymograph construction from coronary angiography.} Spatiotemporal representation of contrast transport enables velocity inference. (a) Sequential fluoroscopy frames show progressive contrast dye advancement through the left anterior descending (LAD) coronary artery. Vessel centerline tracking identifies corresponding anatomical positions across frames. (b) Stacking intensity profiles along the centerline over time constructs a kymograph where the x-axis represents distance along the vessel ($s$) and the y-axis represents time ($t$). Diagonal patterns in the kymograph reflect contrast propagation, with slope $\frac{\Delta s}{\Delta t}$ approximating local blood velocity $u$. This representation transforms the inverse problem of velocity estimation from fluoroscopy into a tractable physics-informed learning task governed by advection-diffusion dynamics.}
    \label{fig:placeholder}
\end{figure}

\section*{Materials and Methods}

\subsection*{Overview of the flow estimation framework}

The goal of the proposed framework is to estimate the CFR from routine coronary angiography while providing an explicit measure of confidence in each prediction. CFR is defined clinically as the ratio of blood flow during pharmacologically induced hyperemia to that at rest and is a key marker of coronary microvascular function.

To achieve this, we model how injected contrast material moves and spreads along the coronary artery over time. We focus on the left anterior descending (LAD) artery as this is the most common artery studied in coronary microvascular dysfunction testing in routine clinical practice \cite{muroya_coronary_2023}. Because coronary flow is predominantly longitudinal over the imaged segment of the LAD, contrast transport can be approximated as a one-dimensional process along the centerline of the vessel (Fig.~\ref{fig:PUNCH_pipeline}). This approximation is widely used in indicator-dilution theory and enables tractable inference from standard fluoroscopic images without requiring three-dimensional vessel reconstruction \cite{zierlerTheoreticalBasisIndicatorDilution1962,zierlerIndicatorDilutionMethods2000}. 

Clinical angiograms often exhibit noise, motion artifacts, variable contrast injection, and incomplete vessel opacification. To account for these factors, the framework explicitly models uncertainty during inference. A shared latent variable of dimension $Z_{dim}$, denoted $\mathbf{z}$, is introduced to represent unobserved sources of data degradation and physiological variability. Conceptually, this latent variable allows the model to express lower confidence when the observed image data are ambiguous or degraded, and higher confidence when the data are clear and consistent with physical expectations \cite{kingmaAutoEncodingVariationalBayes2022, rezendeStochasticBackpropagationApproximate2014}.

From each angiographic sequence, the model estimates three physically significant quantities:

(i) contrast intensity along the vessel as a function of space coordinate ($s$) and time ($t$), $I(s,t;\mathbf{z})$;  

(ii) cross-sectionally averaged and time-averaged axial blood velocity r, $u(s;\mathbf{z})$; and 

(iii) an effective dispersion coefficient, $D(s,t;\mathbf{z})$, which captures contrast spread due to velocity variations, mixing, and unresolved three-dimensional effects.\\
Together, these quantities describe how contrast propagates through the coronary artery and form the basis for estimating flow reserve, all three are conditioned on the latent vector, $\mathbf{z}$. All three are represented using neural networks that are trained jointly under physical constraints. These networks are optimized in such a way that the predicted contrast dynamics satisfies the governing laws of contrast transport while remaining consistent with the observed angiographic intensities.

To derive the law of contrast transport, we introduce another variable $A(s,t)$ to represent the cross-sectional area of the LAD.  (For readability, we omit the variables each quantity depends on). By conserving the mass of blood and contrast fluid in an infinitesimal section of LAD, we arrive at equations:
\begin{equation} \label{transport}
\frac{\partial A}{\partial t} + \frac{\partial}{\partial s}\left(uA \right) = 0~~,~~\frac{\partial }{\partial t}\left(IA\right) + \frac{\partial}{\partial s}\left(u I A\right) = \frac{\partial}{\partial s} \left(A D \frac{\partial I}{\partial s}\right)~.
\end{equation}
Combining the two equations leads to a simplified statement of contrast agent conservation:
\begin{equation} \label{transport}
\frac{\partial I}{\partial t} + u \frac{\partial I}{\partial s} = \frac{1}{A}\frac{\partial (AD)}{\partial s} \frac{\partial I}{\partial s} + D\frac{\partial^2 I}{\partial s^2}~.
\end{equation}
Now, the effective P\'eclet number for contrast agent transported within the LAD is high: in a typical fit of our model to experimental data, we find $P\acute{e}\equiv \dfrac{uL}{D}\sim 4000$, based on a typical vessel length, $L\sim0.1\,$m, dispersion coefficient $D\sim10^{-6}$ m$^2$/s and blood flow velocity $u\sim0.04$ m/s, meaning that the advective terms on the left hand side of the equation dominate over diffusion, which is consistent with imaging data showing a uniform bolus of contrast agent advancing through the vessel. The only place where diffusion is strong enough to balance advection is in (short)  $\sim d/P\acute{e}^{1/2}$ length regions around the leading and trailing fronts of the bolus, where $d$ is the vessel diameter. Because gradients in $I$ are stronger than gradients of $A$ or $D$, within these short regions we may at leading order neglect the first term on the right hand side, whereas, away from the fronts, both diffusion terms are negligible. Hence we arrive at a uniformly valid approximation for the one-dimensional advection--diffusion of contrast agent through the vessel:
\begin{equation} \label{advection-diffusion}
\frac{\partial I}{\partial t}
+ u \frac{\partial I}{\partial s}
- D \frac{\partial^2 I}{\partial s^2}
= 0,
\end{equation}
Here, the diffusion term provides \textit{viscosity regularization} \cite{evansPartialDifferentialEquations2010} of sharp fronts of transported contrast agent.  

Model training is formulated as a multi-objective optimization problem combining three terms:
\begin{equation}
\mathcal{L} = \mathcal{L}_{\text{data}} + \mathcal{L}_{\text{PDE}} + \lambda_{\text{KL}}\, \mathrm{KL}\bigl(q_\phi(\mathbf{z}) \| p(\mathbf{z})\bigr),
\end{equation}
where $\mathcal{L}_{\text{data}}$ enforces agreement with observed angiographic intensities, $\mathcal{L}_{\text{PDE}}$ enforces physical consistency, and the final term regularizes the uncertainty representation. This regularization is a Kullback-Leibler regularization \cite{bishop2006prml} term which prevents overfitting while allowing uncertainty to increase adaptively in regions of poor data quality.

At inference time, uncertainty is propagated by sampling $\mathbf{z}$ from the learned posterior ($N = 100$ samples), producing a distribution of predicted velocity and dispersion fields \cite{galDropoutBayesianApproximation2016, kingmaAutoEncodingVariationalBayes2022}. CFR is computed for each sample as the ratio of spatially averaged velocity during hyperemia to that at rest,
\begin{equation}
\mathrm{CFR} =
\frac{\bar{u}_{\mathrm{hyper}}}{\bar{u}_{\mathrm{rest}}}
=
\frac{\frac{1}{L}\int_0^L u_{\mathrm{hyper}}(s;\mathbf{z})\,\mathrm{d}s}{
\frac{1}{L}\int_0^L u_{\mathrm{rest}}(s;\mathbf{z})\,\mathrm{d}s},
\end{equation}
where $L$ denotes the length of the vessel. Final CFR estimates are reported as the mean and 95\% confidence intervals of the resulting distribution, providing both a point estimate and an explicit measure of confidence.
\begin{figure}[!h]
    \centering
    \includegraphics[width=\linewidth]{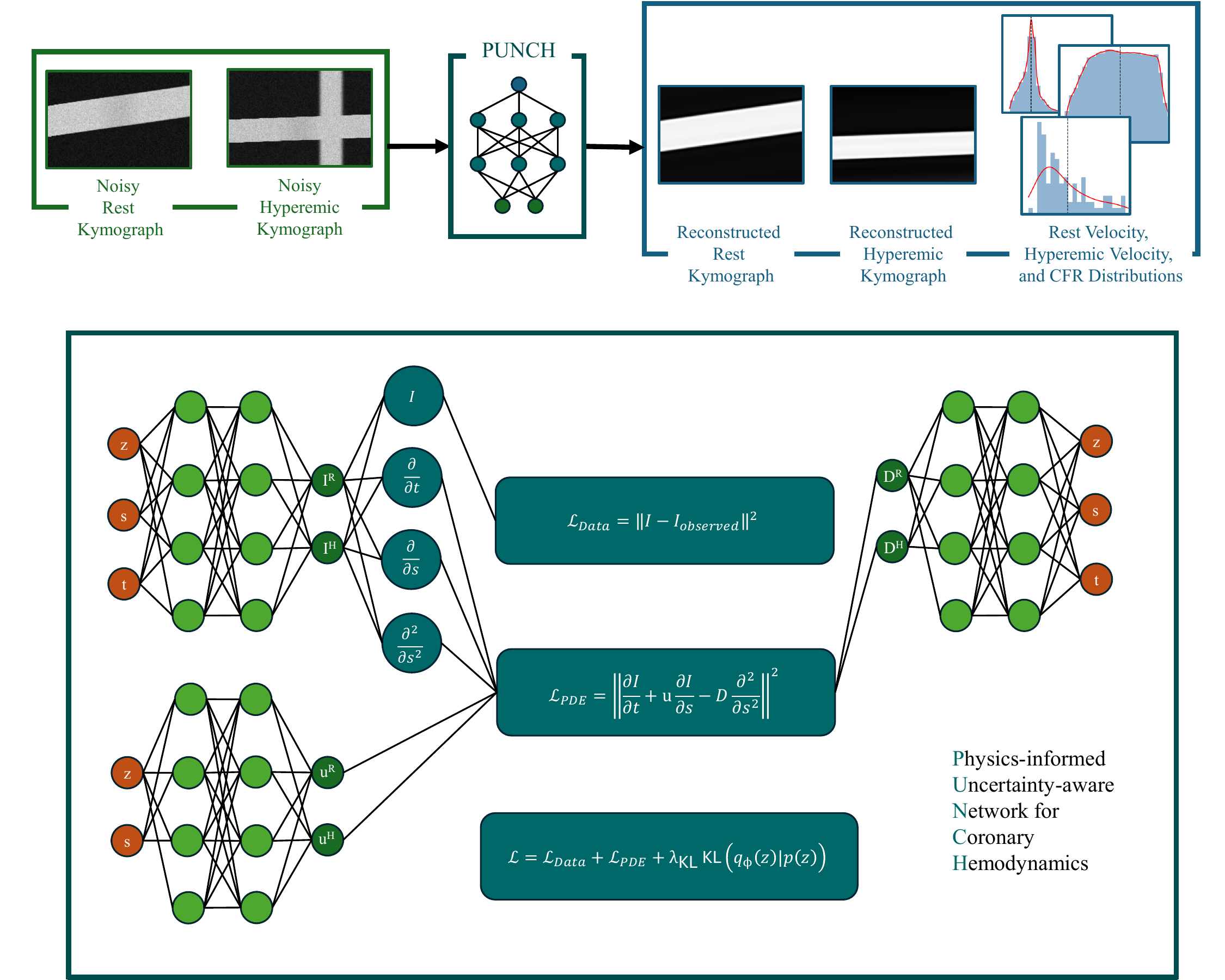}
    \caption{\textbf{Physics-informed neural networks with variational inference enable joint rest-hyperemia modeling.} The framework consists of three subnetworks per physiological state: intensity predictor $I(s,t)$, velocity predictor $u(s)$, and dispersion predictor $D(s,t)$. Rest (subscript R) and hyperemia (subscript H) branches share a common latent variable $z \in \mathbb{R}^{d_z}$ sampled from a learned variational posterior $q_\phi(z)$, ensuring coherent uncertainty propagation between states. Training optimizes a composite loss combining data fidelity to observed noisy kymographs ($\mathcal{L}_{\text{Data}}$), physics-based residuals enforcing the advection-diffusion PDE ($\mathcal{L}_{\text{PDE}}$), and Kullback--Leibler regularization ($\lambda_{\text{KL}} \cdot \text{KL}$). Automatic differentiation computes spatial and temporal derivatives for physics enforcement. At inference, Monte Carlo sampling from the posterior generates distributions over velocity fields and CFR, yielding probabilistic estimates with model-predicted confidence intervals.}
    \label{fig:architecture}
\end{figure}

\subsection*{Network Architecture}

The proposed framework consists of three physics-informed subnetworks that model contrast intensity, axial velocity, and effective dispersion, respectively. Each subnetwork is implemented as a fully connected multilayer perceptron with hyperbolic tangent activations. The intensity network takes spatial position, time, and a shared latent variable as input, whereas the velocity network depends on spatial position and the latent variable, and the dispersion network depends on spatial position, time, and the latent variable. Compact architectures were sufficient for stable convergence, with the intensity network comprising three hidden layers of 64 units, and the velocity and dispersion networks comprising two hidden layers of 32 units each.

To enforce physical plausibility, velocity and dispersion outputs are explicitly bounded through sigmoid-based parameterizations. Velocity predictions are constrained to physiologically realistic ranges after normalization, while dispersion coefficients are restricted to a predefined interval consistent with prior estimates of contrast transport in the coronary arteries. These bounds are applied identically to both the rest state and the hyperemic state.

Rest and hyperemic conditions are modeled using a dual-branch architecture in which each physiological state is represented by an independent set of subnetworks for intensity, velocity, and dispersion, while sharing a common latent variable $\mathbf{z}$ (Fig.~\ref{fig:architecture}). This shared latent representation is inferred via variational inference and captures global, unobserved factors such as image quality, contrast injection variability, and acquisition noise. By coupling the two physiological states through the same latent variable, the uncertainty is coherently propagated between rest and hyperemia, ensuring that the quality of degraded data in one condition induces correlated uncertainty in the other and producing physiologically consistent confidence intervals for CFR.

The latent variable is modeled using a Gaussian variational posterior with learned mean and variance, regularized to a specified prior via a Kullback–Leibler divergence penalty. Sampling from this posterior during training enables uncertainty-aware estimation of flow-related quantities without requiring ground-truth velocity supervision. All subnetworks are trained jointly by minimizing a composite loss comprising physics-based residuals of the advection–diffusion equation, data fidelity terms on observed angiographic intensities, and the variational regularization term.

Rest and hyperemic angiograms were acquired in matched right anterior oblique–cranial projections and point correspondence between frames with a single video was established using point-tracking algorithms (LocoTrack \cite{choLocalAllPairCorrespondence2024}), allowing consistent spatial parameterization across physiological states.

\subsection*{Uncertainty Quantification}

During inference, $N=100$ Monte Carlo samples of $\mathbf{z}$ were drawn to approximate the posterior over flow fields. For each sample $i$, $\text{CFR}_i = \bar{u}_{\mathrm{hyper}}^{(i)} / \bar{u}_{\mathrm{rest}}^{(i)}$ was computed, and the mean and percentile-based 95\% confidence intervals were reported. 
\subsection*{Statistical Analysis}

The correlation between predicted and invasively measured CFR was assessed using Spearman rank correlation. Agreement between methods was evaluated using Bland–Altman analysis \cite{blandStatisticalMethodsAssessing1986} and the concordance correlation coefficient (CCC). Calibration of predicted uncertainty intervals was assessed by computing the fraction of ground-truth (synthetic) or invasive (clinical) CFR values contained within the model’s 95\% confidence intervals. For synthetic datasets where true values were known, the uncertainty–error correlation was calculated to quantify whether higher predicted uncertainty corresponded to greater absolute error. 

\subsection*{Synthetic Data Generation}

\begin{figure}
    \centering
    \includegraphics[width=\linewidth]{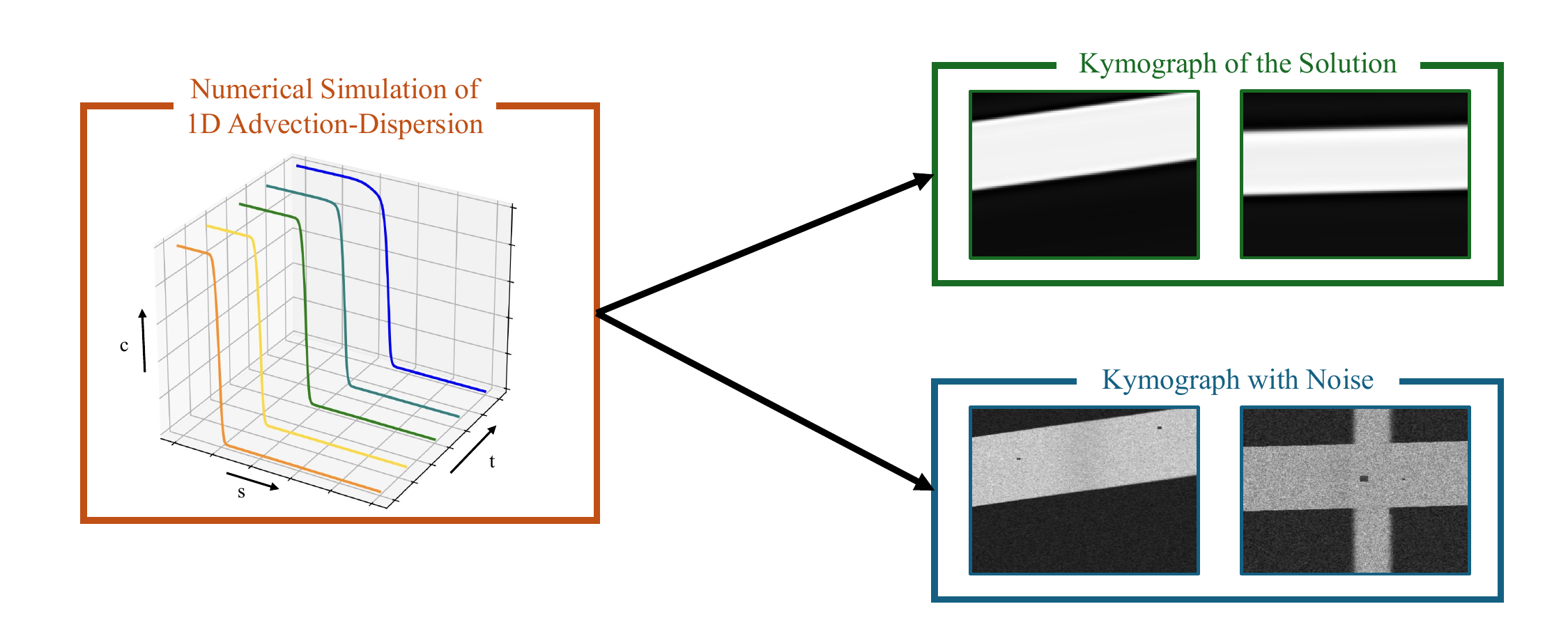}
    \caption{\textbf{Numerical simulation enables inferred CFR values to be compared with known ground truth.} The one-dimensional advection-dispersion equation is solved numerically with randomly generated velocity profiles $u(s)$ and dispersion coefficients $D(s,t)$ to produce noise-free synthetic kymographs. Realistic angiographic noise and artifacts are then applied, including Poisson shot noise, spatial blur, low-frequency intensity variations, vertical banding, localized intensity dropout, and temporal pixelation. This corruption pipeline generates 1,000 synthetic cases spanning mild to severe degradation, enabling quantitative assessment of model accuracy, uncertainty calibration, and robustness to image quality deterioration under controlled conditions where true flow parameters are known.}
    \label{fig:numerical_samlpes}
\end{figure}
To enable controlled evaluation of model accuracy, robustness, and uncertainty calibration under known ground truth, we generated 1,000 synthetic coronary angiogram sequences using a physics-based forward model of contrast transport. The advection-diffusion equation was numerically solved using a method-of-lines framework with high-resolution finite-volume advection and finite-difference diffusion, integrated in time with an explicit Runge–Kutta ODE solver (fig. \ref{fig:numerical_samlpes}). This synthetic framework allowed systematic variation of flow dynamics and image quality while retaining direct access to the true underlying parameters.

The propagation of contrast along the vessel centerline was modeled as a one-dimensional transport process governed by advection and dispersion, reflecting contrast motion with blood flow and its spreading due to velocity heterogeneity and mixing. Specifically, contrast intensity $I(s,t)$ was generated by solving eq. \eqref{advection-diffusion}.
Physiologically plausible velocity profiles were randomly generated for each sequence. Mean resting velocities were sampled between $10$ and $20$ cm/s for typical cases, while 25\% of the sequences were designated as atypical stress-test cases with elevated resting velocities of $20–40$ cm/s. Spatial velocity variations were introduced using smooth, localized modulations along the vessel to mimic focal acceleration or deceleration. Hyperemic flow was simulated by scaling resting velocities using a randomly sampled CFR, drawn from 1.5 to 4.0 for typical cases and 1.0 to 2.0 for low-CFR cases.

The contrast injection was modeled as a finite-duration inlet pulse with a randomized onset time and duration, while the length of the vessel (6–10~cm) and the total simulation time (3–6~s) were independently sampled for rest and hyperemia. The dispersion coefficient was sampled over an order-of-magnitude physiological range to reflect inter-patient variability. All simulated intensity fields were normalized to the unit interval.

To emulate real-world angiographic acquisition, synthetic kymographs were subsequently corrupted with realistic imaging artifacts, including shot noise, spatial blur, low-frequency intensity inhomogeneity, banding artifacts, localized contrast dropout, and temporal downsampling consistent with clinical frame rates. Overall image degradation was modulated by a scalar noise factor spanning mild to severe conditions.

This synthetic dataset enabled rigorous identifiability testing, demonstrating that the proposed framework can uniquely recover ground-truth velocity and dispersion parameters while appropriately inflating predictive uncertainty under degraded image quality. These controlled experiments provide a quantitative benchmark for the accuracy of the model and the calibration of uncertainty prior to clinical validation.

\subsection*{Clinical data acquisition}

All angiographic and invasive measurements were acquired as part of routine clinical care and subsequently used for retrospective analysis from the University of Michigan Health System. Cases included those in whom CFR testing was done for routine clinical indications. Standard coronary angiography was performed in the right anterior oblique–cranial (RAO–CRA) projection at the clinically common frame rates of 10 or 15 frames per second. We used rest and hyperemic angiographic acquisitions from matched projections to minimize foreshortening and motion-related artifacts and to ensure consistent spatial correspondence between physiological states. Briefly, standard acquisition protocols resulted in three bolus injections of room-temperature saline administered at rest to estimate the mean transit time ($\mathrm{Tmn}_{\mathrm{rest}}$). Pharmacological hyperemia was then induced using intravenous adenosine infusion , after which three additional saline bolus injections were performed to measure the hyperemic mean transit time ($\mathrm{Tmn}_{\mathrm{hyp}}$). Coronary flow reserve was computed as the ratio of the average resting to average hyperemic mean transit times \cite{kelshikerCoronaryFlowReserve2022}, as the mean transit time is inversely proportional to the flow rate, consistent with established clinical practice.

\subsection*{Image Processing and Intensity Extraction}

The angiogram of each patient was processed in a standardized workflow:

\begin{enumerate}
\item \textbf{Automated vessel segmentation:} LAD centerlines were extracted using a modified SAM2-UNet architecture \cite{xiongSAM2UNetSegmentAnything2026} with dynamic snake convolution layers \cite{qiDynamicSnakeConvolution2023} which was optimized by minimizing a loss function with binary LAD segmentation masks as targets.  
\item \textbf{Point tracking:} Point correspondence was established between frames within a single video, allowing consistent measurements of image intensity using the LocoTrack algorithm \cite{choLocalAllPairCorrespondence2024}.  
\item \textbf{Contrast isolation:} Multi-scale Hessian filtering was applied to enhance contrast intensity and suppress background structures.  
\item \textbf{Spatial calibration:} DICOM metadata were used to convert pixel coordinates to millimeters.  
\item \textbf{Intensity field construction:} Filtered frames were stacked over time to form a kymograph of $I(s,t)$, where $s$ denotes the arc-length along the centerline of the vessel.
\end{enumerate}

Collocation points for evaluating physics residuals were uniformly sampled along the vessel and across time frames (10 fps) within regions of reliable tracking. The supervised data points corresponded to all spatiotemporal locations with valid Hessian-filtered intensity measurements.

\section*{Results}
\textbf{Clinical interpretation of uncertainty.}
In the proposed framework, each CFR estimate is accompanied by a model-predicted confidence interval that reflects data quality rather than population variability. Narrow confidence intervals indicate that the observed angiographic contrast dynamics are internally consistent and strongly constrain the inferred flow parameters. Conversely, wide intervals signal ambiguity arising from factors such as motion artifacts, incomplete contrast filling, or low signal-to-noise ratio. Clinically, this distinction is critical: low CFR estimates with high confidence may support a diagnosis of coronary microvascular dysfunction, whereas high-uncertainty estimates may instead prompt repeat imaging, alternative projections, or adjunctive physiological testing. By explicitly communicating uncertainty, the framework enables risk-aware interpretation of angiography-derived physiology and reduces the likelihood of false certainty from degraded data.

\subsection*{Validation on synthetic kymographs}

To systematically assess accuracy, calibration, and robustness under controlled conditions, we performed validation on 1,000 synthetic resting angiography kymograph pairs generated with known ground-truth flow dynamics. This synthetic framework allowed direct quantification of the estimation error and uncertainty calibration across a grid of variational hyperparameters, including the prior standard deviation and the strength of the Kullback–Leibler (KL) regularization term.

Throughout the grid search, we observed a clear trade-off between predictive accuracy, uncertainty sharpness, and statistical coverage (Tables~\ref{tab:coverage}–\ref{tab:sharpness}). Increasing the prior standard deviation and the KL weight generally led to higher empirical coverage of the true CFR values (Table~\ref{tab:coverage}), at the cost of increased uncertainty width and reduced sharpness (Table~\ref{tab:sharpness}). For example, coverage improved from approximately 0.67 with low prior variance to values exceeding 0.83 with higher prior variance and stronger KL regularization, indicating more conservative but better-calibrated uncertainty estimates. In contrast, overly strong regularization resulted in inflated uncertainty and degraded point-estimate accuracy, reflected by increased mean absolute error (MAE; Table~\ref{tab:mae}) and root mean square error (RMSE; Table~\ref{tab:rmse}).

Point-estimate accuracy was optimal in an intermediate regime, with MAE values as low as 0.035 (Table~\ref{tab:mae}) and RMSE below 0.10 (Table~\ref{tab:rmse}) for moderate prior variance and KL regularization. In this regime, Deming regression analysis \cite{linnetEvaluationRegressionProcedures1993} demonstrated near-unity slopes (0.96–0.98; Table~\ref{tab:deming_slope}) and small intercepts (Table~\ref{tab:deming_inter}), indicating minimal proportional bias and good agreement with ground truth throughout the dynamic range. The mean bias remained small and consistently negative across all configurations (Table~\ref{tab:bias}), reflecting a mild conservative tendency that increased with stronger regularization.

Beyond aggregate error metrics, we evaluated the diagnostic performance of both posterior means and credible interval bounds. The sensitivity and specificity of the posterior mean exceeded 0.95 in most configurations (Tables~\ref{tab:sens_mean} and~\ref{tab:spec_mean}), while the lower and upper credible bounds maintained high specificity (Tables~\ref{tab:spec_lower} and~\ref{tab:spec_upper}) and sensitivity exceeding 0.90 for moderate regularization strengths (Tables~\ref{tab:sens_lower} and~\ref{tab:sens_upper}). These results indicate that the uncertainty intervals capture the true values with appropriate frequency while remaining narrow enough to preserve clinically meaningful discrimination.

We examined the effect of latent dimensionality in the variational posterior for representative configurations with prior standard deviation $= 10$ and prior standard deviation $= 2$ at $\lambda_{\mathrm{KL}} = 10^{-3}$ (Tables~\ref{tab:zdim_10} and~\ref{tab:zdim_2}). Increasing the latent dimension beyond one to two dimensions resulted in progressively worse performance across multiple metrics, including reduced coverage, increased error, and decreased sensitivity of the lower credible bound. These degradations occurred without compensatory gains in uncertainty sharpness, indicating overparameterization of the latent space. In contrast, low-dimensional latent representations ($Z_{\mathrm{dim}}=1$–2) achieved the best balance between accuracy, calibration, and interval sharpness.

Since real angiographic images may be taken at lower frame rates, we assessed the robustness of flow rate measurement to reduced temporal sampling by varying the effective frame rate of the angiographic sequences for the same representative configurations (Tables~\ref{tab:fps_10} and~\ref{tab:fps_2}). Decreasing the frame rate from 15 to 7.5 frames per second led to a modest reduction in coverage and lower-bound sensitivity, while accuracy metrics (MAE and RMSE) remained largely stable. We similarly analyzed the interplay between accuracy and flow velocity: under hyperemia, flow rates may be fast enough that contrast perfuses the entire LAD over only two angiographic frames, which is too fast for conventional front tracking methods to be accurate. Yet, PUNCH estimates remained accurate for extreme flow velocities (Fig. S7). For low frame rates or fast flow velocities, uncertainty intervals appropriately widened, reflected by increased sharpness, indicating that the model responds to reduced information content by expressing greater uncertainty rather than overconfident predictions. 

Collectively, these results demonstrate that the proposed framework achieves accurate, well-calibrated CFR estimation under controlled synthetic conditions, while explicitly exposing the trade-offs between accuracy and uncertainty across all evaluated hyperparameters. Importantly, the model responds predictably to increased uncertainty in the data-generating process, supporting its use as a reliable uncertainty-aware estimator in downstream clinical applications. Based on this analysis, we selected a prior standard deviation of 2 and $\lambda_{\mathrm{KL}} = 10^{-3}$ for subsequent clinical validation.

\subsection*{Clinical Validation}

Clinical validation was performed in a cohort of 20 patients who underwent a standard invasive CFR assessment using the bolus thermodilution technique \cite{jansenContinuousBolusThermodilutionDerived2023}. For each patient, three repeated measurements were acquired at rest and during pharmacologically induced hyperemia using intravenous adenosine at 140 mcg/kg/min. Invasive CFR was calculated as the ratio of mean hyperemic to mean resting flow values \cite{barbatoValidationCoronaryFlow2004, fearonNovelIndexInvasively2003}. The CFR estimates derived from PUNCH were obtained independently for each case and summarized by their posterior mean and standard deviation.

To quantify experimental uncertainty in the invasive CFR measurements, a nonparametric bootstrap approach was employed. Given three repeated measurements at rest and three during hyperemia, all possible replacement combinations were enumerated for each condition (i.e., $3^3 = 27$ bootstrap samples per state). For each bootstrap realization, CFR was computed as the ratio of the mean hyperemic to the mean resting values. This resulted in $27 \times 27 = 729$ bootstrap CFR estimates per patient, from which the invasive CFR mean and the standard deviation of the sample were computed. This approach captures the uncertainty arising from finite repeated measurements, while preserving the dependence structure inherent to the CFR calculation.

\begin{figure}[!h]
    \centering
    \includegraphics[width=\linewidth]{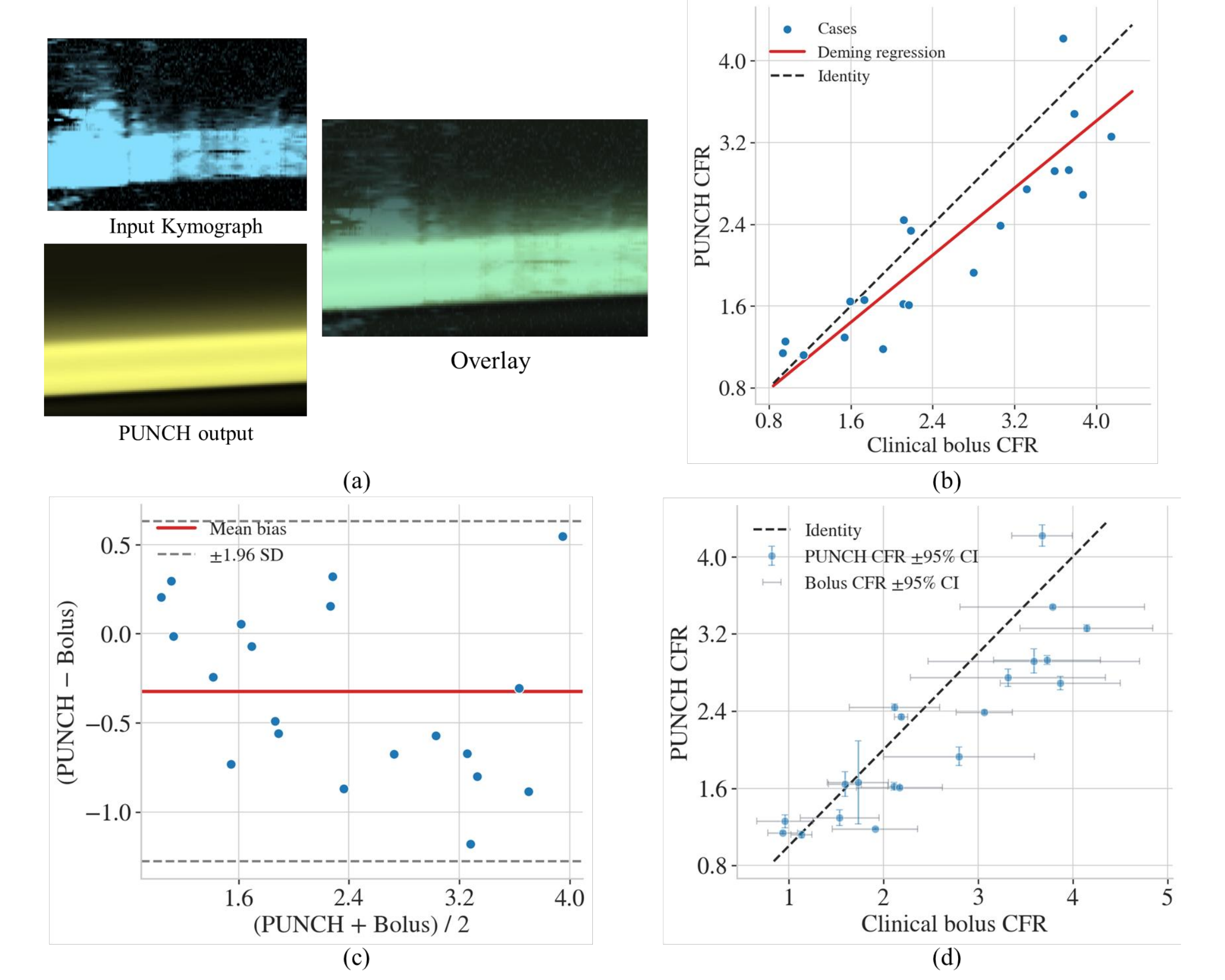}
    \caption{\textbf{PUNCH demonstrates strong agreement with invasive CFR assessment.} (a) Input kymograph (blue) to PUNCH, output kymograph (yellow) and overlay of input and output kymograph (b) Correlation plot comparing PUNCH-derived CFR posterior means against invasive bolus thermodilution measurements in 20 patients shows strong linear association (a Spearman rank correlation of $\rho = 0.89$ ($p < 0.000001$); Deming regression slope $= 0.83$, intercept $= 0.11$).  (c) Bland--Altman analysis reveals mean bias of $-0.31$  (SD $\pm0.48$) with 95\% limits of agreement $[-1.25, 0.63]$, indicating modest systematic underestimation. All 20 cases fall within agreement bounds. (d) Individual patient predictions with 95\% credible intervals (shaded regions) demonstrate consistency between predicted uncertainty and observed invasive values, with narrower PUNCH confidence intervals than intra-patient invasive variability, suggesting improved reproducibility. Error bars represent PUNCH posterior standard deviations and invasive measurement variability across three repeated acquisitions.}
    \label{fig:clinical}
\end{figure}

The PUNCH estimates demonstrated a strong association with invasive measurements, with a Spearman rank correlation of $\rho = 0.89$ ($p < 0.000001$). Ordinary least squares (OLS) regression yielded an $R^2$ of 0.79, indicating that a substantial proportion of the variance in invasive CFR was explained by the PUNCH predictions (Fig.~\ref{fig:clinical}B).

To assess agreement while accounting for measurement error in both modalities, Deming regression was performed. This analysis produced a slope of 0.83 and an intercept of $0.11$ (PUNCH $= 0.83 \times$ Invasive $+ 0.11$), consistent with near-unity proportionality and minimal systematic deviation across the observed CFR range. For comparison, the OLS regression resulted in a slope of 0.75 and an intercept of $0.3$ ($p < 0.000001$). The mean difference between PUNCH and invasive CFR was -0.3132 , while the mean relative difference was -11.43\%.

The agreement was further evaluated by Bland--Altman analysis (Fig.~\ref{fig:clinical}C). The mean difference between PUNCH and invasive CFR was $-0.3132$ , with a standard deviation of 0.48. The corresponding 95\% limits of agreement ranged from $-1.25$ to $0.63$, and all 20 cases fell within these limits. Relative difference analysis revealed a mean bias of $-11.43\%$ (SD $\pm 20.4\%$), with 10 of 20 cases (50\%) within $\pm 20\%$ agreement (Fig.~\ref{fig:clinical}C). Collectively, these findings indicate a modest negative bias, reflecting a tendency toward lower CFR estimates relative to invasive measurements. It should be emphasized that bolus thermodilution has been reported to yield systematically higher CFR values than continuous thermodilution, although both methods are, in principle, measuring identical physiological variables \cite{gallinoroReproducibilityBolusContinuous2023,jansenContinuousBolusThermodilutionDerived2023,fawazComparisonBolusContinuous2024}. The methodological tendency of bolus thermodilution toward higher CFR may contribute to the modest negative bias observed between PUNCH estimates and bolus thermodilution CFR values. 

In addition to point estimates, the PUNCH framework provides posterior uncertainty predicted by the model for each CFR estimate. In Fig.~\ref{fig:clinical}D, PUNCH posterior means and 95\% confidence intervals are shown alongside invasive CFR means and bootstrap-derived standard deviations. The overlap between PUNCH credible intervals and invasive bootstrap uncertainty supports the internal calibration of the probabilistic model and demonstrates the concordance between predicted and experimentally observed variability. These results highlight the potential value of uncertainty-aware, non-invasive CFR estimation in clinical and research settings.

\section*{Discussion}

We present a physics-informed, uncertainty-aware framework for non-invasive diagnosis of coronary microvascular dysfunction (CMD) that transforms routine angiography into a functional assessment tool. Our approach integrates (1) physics-based learning directly from unlabeled angiographic sequences with per-patient inference; (2) variational inference that produces calibrated uncertainty estimates tied to image quality; and (3) a dual-branch architecture that jointly models resting and hyperemic states to produce physiologically consistent CFR confidence intervals. By extracting quantitative physiology from standard-of-care imaging, this framework addresses a critical unmet need in cardiovascular diagnostics. Importantly, the complete analysis pipeline is executed in approximately three minutes per patient on a single NVIDIA A40 GPU, without the need for prior population-level training, allowing rapid and on-demand physiological assessment directly at the point of care.

\textbf{Disease Burden and Diagnostic Gaps.} Coronary microvascular dysfunction represents a major public health challenge affecting millions of patients worldwide. CMD is associated with a nearly 4-fold increase in all-cause mortality 
and a 5-fold increase in major adverse cardiac events 
compared to patients without microvascular disease~\cite{gdowskiAssociationIsolatedCoronary2020}. Patients with CMD experience substantially elevated risks of cardiovascular death, myocardial infarction, hospitalization for heart failure, and stroke~\cite{kelshikerCoronaryFlowReserve2022,taquetiCoronaryMicrovascularDisease2018}. With approximately 30--40\% of non-obstructive coronary artery disease patients demonstrating evidence of CMD after systematic evaluation~\cite{kunadianEAPCIExpertConsensus2021}, and 3--4 million patients in this category in the United States alone, under-diagnosis of CMD is an enormous public health problem ~\cite{baireymerzIschemiaNoObstructive2017,tjoeCoronaryMicrovascularDysfunction2021}.

However, despite its disease burden, CMD remains deeply underdiagnosed. Approximately 4 million coronary angiograms are performed annually in Europe and the United States combined~\cite{bestemirAnalysisCoronaryAngiography2023,fordStratifiedMedicalTherapy2018}, representing one of the most common invasive cardiovascular procedures worldwide. By contrast, specialized invasive CMD assessments using bolus or continuous thermodilution are performed in only a few thousands of patients annually, concentrated in a small number of specialized centers with dedicated expertise~\cite{barbatoValidationCoronaryFlow2004}. The striking disparity between millions of angiograms and thousands of functional microvascular assessments leaves the vast majority of patients without a definitive functional diagnosis. Standard angiography, while providing excellent anatomical resolution of the epicardial vessels, does not provide direct information about microvascular function, perpetuating diagnostic uncertainty in a high-risk population.

\textbf{Transformative Clinical Potential.} By unlocking physiological information already encoded in routine angiographic imaging, our method could extend CMD assessment to millions of patients  without requiring additional specialized equipment, separate procedures, or advanced imaging modalities. This capability is particularly consequential in low- and middle-income regions where coronary angiography is increasingly available, but access to advanced  microvascular testing or physiological imaging remains highly limited, creating a substantial global diagnostic gap. By converting an already ubiquitous procedure into a quantitative functional assessment, the proposed framework has the potential to democratize access to CMD evaluation in diverse clinical and resource settings. CMD is also disproportionately underdiagnosed in women around the world, who frequently present with angina in the absence of obstructive coronary disease and are less likely to undergo invasive physiological tests. The explicit modeling of uncertainty is critical for safe clinical deployment: confident low CFR predictions may inform immediate therapeutic decisions, whereas appropriately inflated uncertainty intervals in the setting of poor image quality can trigger repeat imaging or adjunctive testing, reducing both false diagnoses and unnecessary interventions. This uncertainty-aware approach embodies the principle of ``honest AI,'' where the system's confidence directly reflects data quality and diagnostic reliability at the point of care.

\textbf{Comparison to Existing Paradigms.} Computational fluid dynamics (CFD)-based methods require high-resolution CT imaging followed by three-dimensional vessel reconstruction, and extensive computation (often hours per patient), fundamentally limiting its clinical scalability~\cite{taylorComputationalFluidDynamics2013,morrisComputationalFluidDynamics2016}. Supervised machine learning approaches require large labeled datasets that simply do not exist for challenging phenotypes such as CMD, where invasive ground truth measurements are available only in small registries~\cite{ituMachinelearningApproachComputation2016,liAutomaticCoronaryArtery2022}. In contrast, our framework learns directly from routinely acquired angiograms without labeled flow data, computes patient-specific solutions in minutes, and provides probabilistic estimates with quantified uncertainty; capabilities absent in prior approaches. The per-patient training paradigm eliminates concerns about generalization between heterogeneous populations or data leakage between training and test sets. In particular, our predicted CFR confidence intervals were narrower than the intra-patient variability observed in repeated invasive thermodilution measurements (median standard deviation approximately 12\%)~\cite{barbatoValidationCoronaryFlow2004}, indicating both precision and potential for improved reproducibility relative to the current invasive methods. 

\textbf{Limitations.} The current framework assumes quasi-one-dimensional flow along the vessel centerline; complex anatomies involving severe bifurcations, vessel overlap, or extreme tortuosity may violate this approximation and reduce inference fidelity. Systematic characterization of these failure modes, together with extension to multi-branch coronary geometries, represents an important direction for future methodological development. Beyond model generalization, larger multi-center studies will be required to establish clinically actionable diagnostic thresholds, evaluate robustness across diverse patient populations and anatomical variants, and relate PUNCH-derived physiological estimates to downstream clinical decision-making and outcomes. In this work, we focus on the LAD as it is the most common artery studied in coronary microvascular dysfunction testing in routine clinical practice. We do not anticipate technical challenges to extending the method to the right coronary arteries. 

\textbf{Broader Implications.} This work demonstrates that embedding known physical laws into machine learning with rigorous uncertainty quantification can address a longstanding barrier to clinical AI adoption: the inability to distinguish confident predictions from unreliable extrapolations. By converting a routinely acquired imaging procedure into a quantitative diagnostic tool, the proposed framework raises broader questions about how much latent physiological information is already encoded in standard clinical data but remains untapped. When does a common imaging modality become a functional test, and what level of uncertainty is sufficient to support clinical decision-making rather than exploratory analysis?

The per-patient training paradigm further challenges the prevailing assumption that large, multi-center labeled datasets are a prerequisite for clinical translation, suggesting an alternative route for biomedical AI deployment when governing physics is well understood but ground-truth measurements are scarce. More broadly, this framework invites investigation into which other medical imaging–based inverse problems—such as perfusion imaging, tracer kinetics, or dynamic contrast studies across vascular beds—may be similarly reframed as physics-constrained inference tasks, transforming existing clinical workflows without requiring new hardware, new acquisitions, or large-scale annotation efforts.





\section*{Competing Interests}

Sukirt Thakur, Yang Zhou, Dmitry Yu. Isaev, Srinivas Paruchuri, and Scott Burger are employees of and hold equity interests in AngioInsight, Inc., a company developing artificial intelligence solutions in cardiac health. Marcus Roper and Maziar Raissi are hold an equity interest in AngioInsight, Inc. as consultants. Brahmajee K. Nallamothu and C. Alberto Figueroa are employed by the University of Michigan and hold an equity interest in AngioInsight, Inc. as consultants and co-founders.

\section*{Data and Code Availability}

Clinical data cannot be shared publicly due to patient privacy but are available from the corresponding author upon reasonable request and with appropriate institutional approvals. Synthetic validation data and model code can be made available upon reasonable request and with appropriate institutional approvals.

\bibliographystyle{unsrt}   
\bibliography{references}

\section*{Supplementary Materials}
\label{sec:supplementary}

\subsection*{Training Protocol}

For each patient, rest and hyperemic angiograms were modeled jointly using a dedicated dual-branch physics-informed neural network, with no parameter sharing across patients to prevent data leakage. All subnetworks and latent variables were optimized end-to-end using the Adam optimizer with an initial learning rate of $10^{-3}$. A cosine annealing learning rate schedule was employed over 10{,}000 training epochs.

At each iteration, collocation points for enforcing the governing advection–diffusion equation were sampled independently for rest and hyperemic states from fine spatiotemporal grids, with 2{,}000 points per state. Supervised data fidelity losses were computed using 2{,}000 randomly sampled observed intensity points from each angiogram. All spatial and temporal derivatives required for the physics-based residuals were computed using automatic differentiation.

Uncertainty is modeled through a variational latent variable framework. The posterior distribution over $\mathbf{z}$ is defined as
$q_\phi(\mathbf{z}) = \mathcal{N}(\boldsymbol{\mu}, \mathrm{diag}(\boldsymbol{\sigma}^2))$
and is regularized toward a standard normal prior using a Kullback--Leibler divergence term. This regularization prevents overfitting while allowing the model to adaptively increase uncertainty in regions of poor data quality. The weighting factor was selected to balance data fidelity and uncertainty regularization.

Training and inference with 100 Monte Carlo samples and CFR computation required about ~3 minutes per patient on a single NVIDIA A40 GPU.

\begin{table}[h!]
\centering
\caption{Grid search - Coverage}
\label{tab:coverage}
\begin{tabular}{c|S[table-format=1.3]S[table-format=1.3]S[table-format=1.3]}
\toprule
Prior Std dev & {KL = 1e-4} & {KL = 1e-3} & {KL = 1e-2} \\
\midrule
2  & 0.674 & 0.671 & 0.685 \\
5  & 0.717 & 0.749 & 0.837 \\
10 & 0.719 & 0.838 & 0.867 \\
\bottomrule
\end{tabular}
\end{table}

\begin{table}[h!]
\centering
\caption{Grid search - MAE}
\label{tab:mae}
\begin{tabular}{c|S[table-format=1.3]S[table-format=1.3]S[table-format=1.3]}
\toprule
Prior Std dev & {KL = 1e-4} & {KL = 1e-3} & {KL = 1e-2} \\
\midrule
2  & 0.0651  & 0.0345 & 0.0526 \\
5  & 0.0393  & 0.0465 & 0.0739 \\
10 & 0.0366  & 0.0457 & 0.0923 \\
\bottomrule
\end{tabular}
\end{table}

\begin{table}[h!]
\centering
\caption{Grid search - RMSE}
\label{tab:rmse}
\begin{tabular}{c|S[table-format=1.3]S[table-format=1.3]S[table-format=1.3]}
\toprule
Prior Std dev & {KL = 1e-4} & {KL = 1e-3} & {KL = 1e-2} \\
\midrule
2  & 0.0725 & 0.0701 & 0.3809 \\
5  & 0.1149 & 0.1122 & 0.1415 \\
10 & 0.0989 & 0.0948 & 0.1745 \\
\bottomrule
\end{tabular}
\end{table}

\begin{table}[h!]
\centering
\caption{Grid search - Sharpness}
\label{tab:sharpness}
\begin{tabular}{c|S[table-format=1.3]S[table-format=1.3]S[table-format=1.3]}
\toprule
Prior Std dev & {KL = 1e-4} & {KL = 1e-3} & {KL = 1e-2} \\
\midrule
2  & 0.0797 & 0.0695 & 0.0766 \\
5  & 0.0864 & 0.1139 & 0.223 \\
10 & 0.0831 & 0.148 & 0.2784 \\
\bottomrule
\end{tabular}
\end{table}

\begin{table}[h!]
\centering
\caption{Grid search - Deming regression slope}
\label{tab:deming_slope}
\begin{tabular}{c|S[table-format=1.3]S[table-format=1.3]S[table-format=1.3]}
\toprule
Prior Std dev & {KL = 1e-4} & {KL = 1e-3} & {KL = 1e-2} \\
\midrule
2  & 0.978 & 0.977 & 0.976 \\
5  & 0.974  & 0.965 & 0.940 \\
10 & 0.978 & 0.965 & 0.925 \\
\bottomrule
\end{tabular}
\end{table}

\begin{table}[h!]
\centering
\caption{Grid search - Deming regression intercept}
\label{tab:deming_inter}
\begin{tabular}{c|S[table-format=1.3]S[table-format=1.3]S[table-format=1.3]}
\toprule
Prior Std dev & {KL = 1e-4} & {KL = 1e-3} & {KL = 1e-2} \\
\midrule
2  & 0.023 & 0.026 & 0.028 \\
5  & 0.031  & 0.043 & 0.073 \\
10 & 0.024 & 0.041 & 0.090 \\
\bottomrule
\end{tabular}
\end{table}

\begin{table}[h!]
\centering
\caption{Grid search - Bias (Mean - True)}
\label{tab:bias}
\begin{tabular}{c|S[table-format=1.3]S[table-format=1.3]S[table-format=1.3]}
\toprule
Prior Std dev & {KL = 1e-4} & {KL = 1e-3} & {KL = 1e-2} \\
\midrule
2  & 0.0028 & -0.0276 & -0.0094 \\
5  & -0.0247  & -0.0359 & -0.0657 \\
10 & -0.0255 & -0.0405 & -0.0852 \\
\bottomrule
\end{tabular}
\end{table}

\begin{table}[h!]
\centering
\caption{Grid search - Sensitivity of mean}
\label{tab:sens_mean}
\begin{tabular}{c|S[table-format=1.3]S[table-format=1.3]S[table-format=1.3]}
\toprule
Prior Std dev & {KL = 1e-4} & {KL = 1e-3} & {KL = 1e-2} \\
\midrule
2  & 0.977 & 0.974 & 0.979 \\
5  & 0.979  & 0.974 & 0.959 \\
10 & 0.969 & 0.974 & 0.948 \\
\bottomrule
\end{tabular}
\end{table}

\begin{table}[h!]
\centering
\caption{Grid search - Specificity of mean}
\label{tab:spec_mean}
\begin{tabular}{c|S[table-format=1.3]S[table-format=1.3]S[table-format=1.3]}
\toprule
Prior Std dev & {KL = 1e-4} & {KL = 1e-3} & {KL = 1e-2} \\
\midrule
2  & 0.997 & 0.997 & 0.997 \\
5  & 0.995  & 0.997 & 0.997 \\
10 & 0.997 & 1.00 & 0.997 \\
\bottomrule
\end{tabular}
\end{table}

\begin{table}[h!]
\centering
\caption{Grid search - Sensitivity of lower bound}
\label{tab:sens_lower}
\begin{tabular}{c|S[table-format=1.3]S[table-format=1.3]S[table-format=1.3]}
\toprule
Prior Std dev & {KL = 1e-4} & {KL = 1e-3} & {KL = 1e-2} \\
\midrule
2  & 0.938 & 0.951 & 0.943 \\
5  & 0.941  & 0.928 & 0.845 \\
10 & 0.943 & 0.904 & 0.769 \\
\bottomrule
\end{tabular}
\end{table}

\begin{table}[h!]
\centering
\caption{Grid search - Specificity of lower bound}
\label{tab:spec_lower}
\begin{tabular}{c|S[table-format=1.3]S[table-format=1.3]S[table-format=1.3]}
\toprule
Prior Std dev & {KL = 1e-4} & {KL = 1e-3} & {KL = 1e-2} \\
\midrule
2  & 0.998 & 0.998 & 1.000 \\
5  & 1.000  & 1.000 & 1.000 \\
10 & 0.997 & 1.000 & 1.000 \\
\bottomrule
\end{tabular}
\end{table}

\begin{table}[h!]
\centering
\caption{Grid search - Sensitivity of upper bound}
\label{tab:sens_upper}
\begin{tabular}{c|S[table-format=1.3]S[table-format=1.3]S[table-format=1.3]}
\toprule
Prior Std dev & {KL = 1e-4} & {KL = 1e-3} & {KL = 1e-2} \\
\midrule
2  & 0.992 & 0.992 & 0.990 \\
5  & 0.997  & 0.984 & 0.995 \\
10 & 0.990 & 0.995 & 0.997 \\
\bottomrule
\end{tabular}
\end{table}

\begin{table}[h!]
\centering
\caption{Grid search - Specificity of upper bound}
\label{tab:spec_upper}
\begin{tabular}{c|S[table-format=1.3]S[table-format=1.3]S[table-format=1.3]}
\toprule
Prior Std dev & {KL = 1e-4} & {KL = 1e-3} & {KL = 1e-2} \\
\midrule
2  & 0.987 & 0.987 & 0.988 \\
5  & 0.990  & 0.985 & 0.970 \\
10 & 0.988 & 0.977 & 0.954 \\
\bottomrule
\end{tabular}
\end{table}

\begin{table}[h!]
\centering
\caption{Grid search - Zdim for std 10 and KL 1e-3}
\label{tab:zdim_10}
\begin{tabular}{c|S[table-format=1.3]S[table-format=1.3]S[table-format=1.3]S[table-format=1.3]}
\toprule
Metric & {$Z_{dim} = 1$} & {$Z_{dim} = 2$} & {$Z_{dim} = 4$} & {$Z_{dim} = 8$} \\
\midrule
Coverage  & 0.838 & 0.839 & 0.686 & 0.484 \\
MAE  & 0.0457  & 0.0675 & 0.1144 & 0.1747\\
RMSE & 0.0948 & 0.1319 & 0.2052 & 0.2918 \\
Sharpness & 0.148 & 0.1924 & 0.2037 & 0.1706 \\
Sensitivity (Lower) & 0.974 & 0.873 & 0.803 & 0.710 \\
Specificity (Lower) & 1.000 & 1.000 & 1.000 & 1.000 \\
\bottomrule
\end{tabular}
\end{table}

\begin{table}[h!]
\centering
\caption{Grid search - The effect of frames per seccond for std 10 and KL 1e-3}
\label{tab:fps_10}
\begin{tabular}{c|S[table-format=1.3]S[table-format=1.3]S[table-format=1.3]S[table-format=1.3]}
\toprule
Metric & {$fps=15$} & {$fps=10$} & {$fps=7.5$} \\
\midrule
Coverage  & 0.838 & 0.814 & 0.787 \\
MAE  & 0.0457  & 0.0490 & 0.0463 \\
RMSE & 0.0948 & 0.0952 & 0.0848 \\
Sharpness & 0.148 & 0.1516 & 0.1610 \\
Sensitivity (Lower) & 0.974 & 0.912 & 0.922  \\
Specificity (Lower) & 1.000 & 0.998 & 1.000  \\
\bottomrule
\end{tabular}
\end{table}

\begin{table}[h!]
\centering
\caption{Grid search - Zdim for std 2 and KL 1e-3}
\label{tab:zdim_2}
\begin{tabular}{c|S[table-format=1.3]S[table-format=1.3]S[table-format=1.3]S[table-format=1.3]}
\toprule
Metric & {$Z_{dim} = 1$} & {$Z_{dim} = 2$} & {$Z_{dim} = 4$} & {$Z_{dim} = 8$} \\
\midrule
Coverage  & 0.672 & 0.799 & 0.710 & 0.691 \\
MAE  & 0.0340  & 0.0456 & 0.0803 & 0.0803\\
RMSE & 0.0684 & 0.0949 & 0.1499 & 0.1470 \\
Sharpness & 0.0693 & 0.1179 & 0.1494 & 0.1515 \\
Sensitivity (Lower) & 0.951 & 0.917 & 0.881 & 0.889 \\
Specificity (Lower) & 0.998 & 0.998 & 1.000 & 1.000 \\
\bottomrule
\end{tabular}
\end{table}

\begin{table}[h!]
\centering
\caption{Grid search - The effect of frames per seccond for std 2 and KL 1e-3}
\label{tab:fps_2}
\begin{tabular}{c|S[table-format=1.3]S[table-format=1.3]S[table-format=1.3]S[table-format=1.3]}
\toprule
Metric & {$fps=15$} & {$fps=10$} & {$fps=7.5$} \\
\midrule
Coverage  & 0.672 & 0.680 & 0.674 \\
MAE  & 0.0340  & 0.0837 & 0.086 \\
RMSE & 0.0684 & 0.1489 & 0.1576 \\
Sharpness & 0.0693 & 0.1517 & 0.1548 \\
Sensitivity (Lower) & 0.951 & 0.876 & 0.857  \\
Specificity (Lower) & 0.998 & 1.000 & 1.000  \\
\bottomrule
\end{tabular}
\end{table}

\begin{figure}
    \centering
    \includegraphics[width=\linewidth]{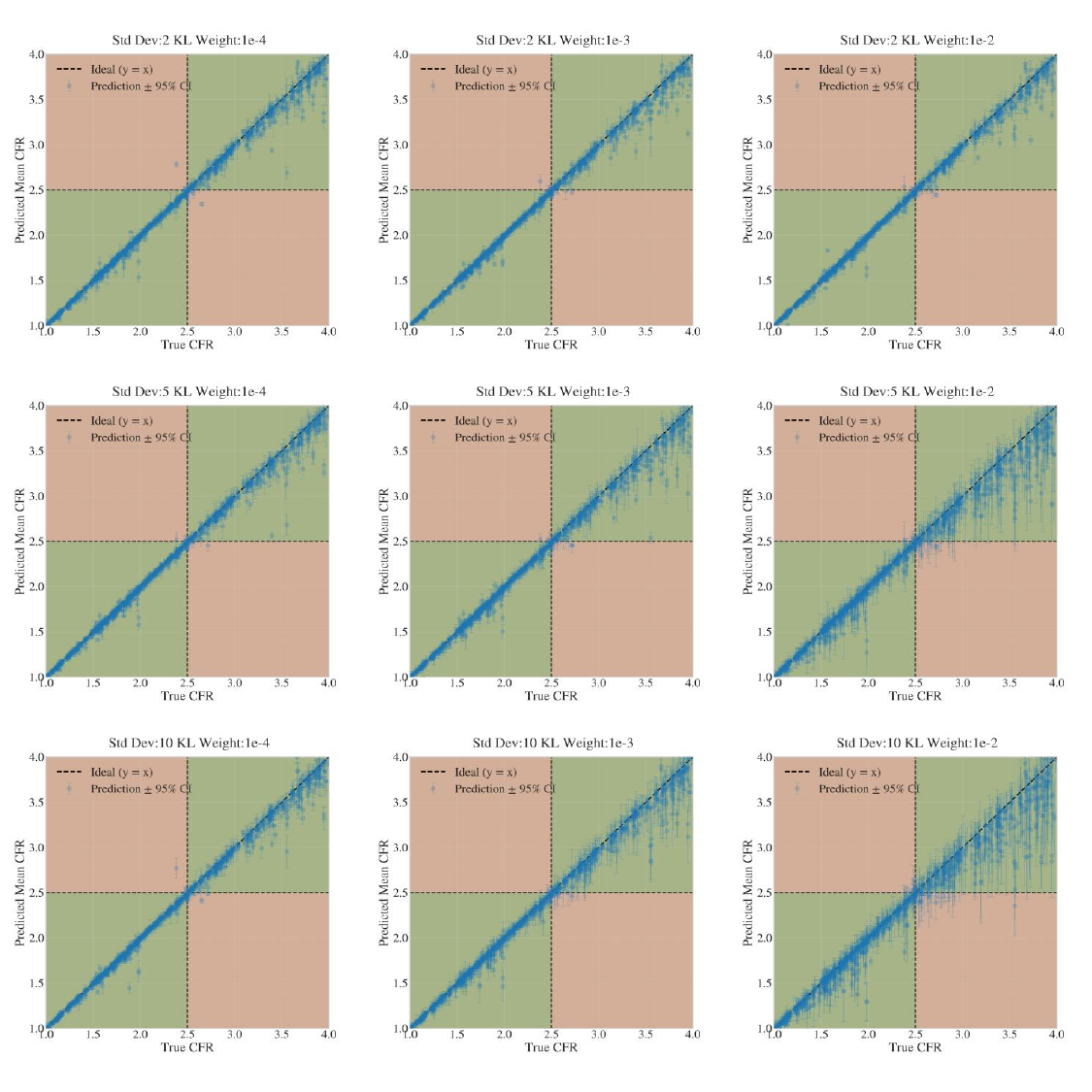}
    \caption{\textbf{Uncertainty-error correlation in synthetic validation.} Model uncertainty reliably reflects estimation error magnitude. Analysis of 1,000 synthetic cases demonstrates strong correspondence between PUNCH-predicted posterior standard deviation and absolute CFR estimation error (Pearson $r = 0.997$, Spearman $\rho = 0.998$). Cases with high image degradation (noise factor 2.5--3.0) exhibit appropriately inflated uncertainty estimates and larger errors, while high-quality cases (noise factor 0.5--1.0) yield precise estimates with narrow confidence intervals. This calibration confirms the framework's capacity to identify degraded data quality and respond with honest uncertainty quantification---a critical requirement for safe clinical deployment. Scatter plot color-codes synthetic cases by noise severity; diagonal reference line indicates perfect uncertainty-error correspondence.}
    \label{fig:placeholder}
\end{figure}

\begin{figure}[!h]
    \centering
    \includegraphics[width=0.75\linewidth]{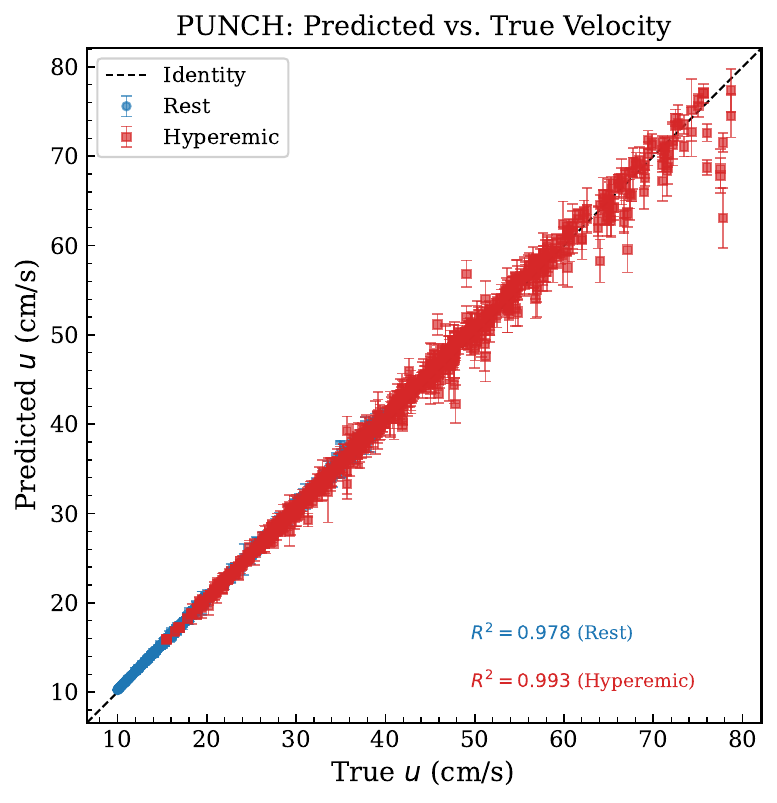}
    \caption{PUNCH accurately recovers blood velocity from synthetic angiograms. Predicted versus true spatially averaged blood velocity for rest (blue circles) and hyperemic (red squares) conditions across 1,000 synthetic cases. Error bars indicate posterior standard deviations from Monte Carlo sampling. PUNCH achieves near-unity agreement with ground truth across the full physiological range, with $R^2 = 0.978$ at rest and $R^2 = 0.993$ during hyperemia. The dashed line denotes the identity. The tighter clustering and higher $R^2$ under hyperemia reflect the greater dynamic range and stronger advective signal at elevated flow velocities, which better constrain the physics-informed inference.}
    \label{fig:fig8}
\end{figure}

Figure~\ref{fig:fig8} demonstrates that PUNCH accurately recovers blood 
velocity throughout the full physiological range under both rest and hyperemic conditions. The predictions closely follow the identity line with $R^2 = 0.978$ at rest and $R^2 = 0.993$ during hyperemia, indicating minimal proportional bias and strong agreement with ground truth. The posterior 
standard deviations, shown as error bars, remain narrow throughout, reflecting well-constrained 
inference under the advection-diffusion physics. The marginally higher $R^2$ under hyperemia is 
consistent with the larger dynamic range and stronger advective signal at elevated flow velocities, 
which provide a richer contrast transport pattern for the network to exploit. Taken together, 
these results confirm that PUNCH resolves the underlying hemodynamic quantities, not only their ratio, with high fidelity, supporting the reliability of the derived CFR estimates reported in the main text.

\end{document}